\documentclass{article}
\usepackage{spconf,amsmath,graphicx,xspace, booktabs, adjustbox}
\usepackage[breaklinks=true,letterpaper=true,colorlinks,bookmarks=false, citecolor=cyan, linkcolor=magenta]{hyperref}
\newcommand{\fare}{FaRE}
\newcommand{\latinphrase}[1]{\textit{#1}}
\newcommand{\etal}{\latinphrase{et~al.}\xspace}

\newcommand{\etc}{\latinphrase{etc.}\xspace}

\title{Open Source Face Recognition Performance Evaluation Package}
\name{Xiang Xu and Ioannis A. Kakadiaris}
\address{Computational Biomedicine Lab \\ University of Houston \\ 4849 Calhoun Road, Houston, TX, 77204}

\begin{document}
%
\maketitle
\begin{abstract}
Biometrics-related research has been accelerated significantly by deep learning technology.
However, there are limited open-source resources to help researchers evaluate their deep learning-based biometrics algorithms efficiently, especially for the face recognition tasks.
In this work, we design and implement a light-weight, maintainable, scalable, generalizable, and extendable face recognition evaluation toolbox named FaRE that supports both online and offline evaluation to provide feedback to algorithm development and accelerate biometrics-related research.
FaRE consists of a set of evaluation metric functions and provides various APIs for commonly-used face recognition datasets including LFW, CFP, UHDB31, and IJB-series datasets, which can be easily extended to include other customized datasets.
The package and the pre-trained baseline models will be released for public academic research use after obtaining university approval.
\end{abstract}
\begin{keywords}
Face Recognition, Evaluation, Toolbox
\end{keywords}
\section{Introduction}
\label{SEC:FaRE-Intro}
\begin{figure}[tb]
    \centering
    \centerline{\includegraphics[width=\linewidth]{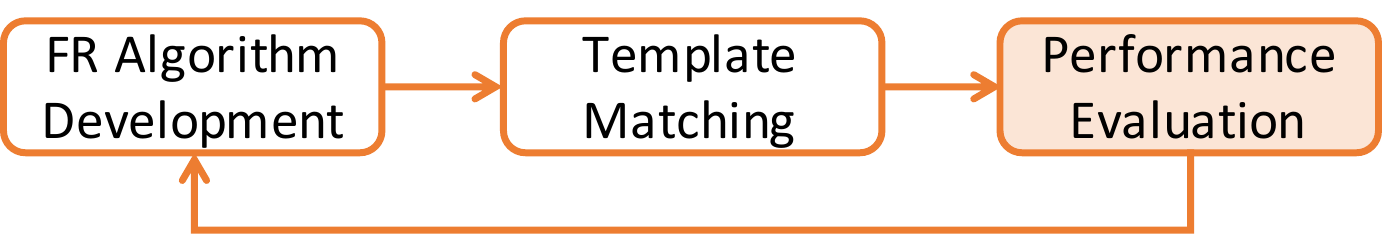}}
    \caption{Depiction of the general FR development pipeline consists of FR algorithm development (template generation), template matching, and performance evaluation. Performance evaluation can provide feedback to algorithm development. Our work fits in the performance evaluation block.}
    \label{FIG:FaRE-Pip}
\end{figure}

Face recognition (FR) is a general concept for the applications of face identification and verification. 
In the scenario of face identification, the system classifies human identity according to a single facial image or a set of images while in the scenario of face verification, a binary decision is made by the computer to decide whether two images belong to the same identity. 
Recently, the deployment of deep neural network has produced impressive results in both face identification and verification tasks. 
Methods such as ArcFace~\cite{Deng_2018_180917} have significantly pushed the frontier of face recognition performance. 
Figure~\ref{FIG:FaRE-Pip} depicts a general face recognition system consists of three stages including enrollment, matching, and evaluation stages.
During the enrollment, the algorithm should generate a template from a facial image or a set of images for each subject.
Most of the current research work is focusing on learning to generate a discriminative template for each identity.
In the matching stage, the distance or similarity score is computed and a decision algorithm determines the identity in an identification application or to accept/reject the person in a verification scenario.
In the evaluation stage, the overall face recognition performance is assessed using quantitative measurements.
However, in the biometrics community, there are limited existing resources that provide a consistent and general measurement for evaluating and developing a deep learning-based face recognition system.
This need becomes more acute, especially for the on-boarding of new researchers.

In this work, we design and implement a light-weight, maintainable, scalable, generalizable, and extendable face recognition evaluation package named FaRE, which can be generalized to the data and algorithms developed from other modalities such as fingerprint, \etc.
The whole package, which is easy to install, is written in Python since most of the current deep learning frameworks support Python language.
We adopt the most commonly-used FR datasets, analyze their metrics, and then generalize an evaluation pipeline to evaluate the performance of FR algorithms.
To support offline evaluation, a file management module is implemented to organize and match the generated template file with meta-data for each dataset. 
To support online evaluation, data loaders are implemented to feed the data to the neural network and generate a template from a facial image or an image set.
The similarity matrix is obtained by computing the similarity of the templates from probe and gallery set based on the evaluating dataset protocol.
Based on the similarity matrix and the ground-truth label provided in datasets, different quantitative measurement functions are used according to the protocols provided in the datasets.
To visualize the quantitative results, comparison figures can be plotted using FaRE.
With our evaluation package, the new datasets and protocols can be easily extended and evaluated.
In addition, new fusion functions can be easily added for set-based FR.
FaRE is being used to validate the biometrics algorithms developed in the UH Computational Biomedicine Lab (CBL).
To provide the pre-trained models for the biometrics community, two face template generators are trained using \mbox{ResNet-101 \cite{He_2016_17161}} and DenseNet-121 \cite{Huang_2017_190102} architectures on VGG-Face2 \cite{Cao_2018_17830} dataset.
The FR experiments including open-set face identification and face verification are performed and are evaluated by FaRE.
In summary, our contributions are two-fold: (i) A light-weighted, maintainable, scalable, generalizable, and extendable FR package is designed and implemented; (ii) Two networks are trained and evaluated on IJB-C, which can serve pre-trained models in other face-related tasks.


\section{Related Work}
\label{SEC:FaRE-LR}
\noindent \textbf{FR Algorithms, Datasets, and Protocols}: 
A significant part of the success of recent FR algorithms can be attributed to the large-scale image collections \cite{Cao_2018_17830} that have become available in the past few years. 
Researchers improve the FR performance by deploying the generative model to generate the frontal images \cite{Tran_2017_17705} or developing new loss functions \cite{Liu_2017_180923, Deng_2018_180917} to learn discriminative face representations.
To evaluate the improvement of unconstrained FR algorithms, several benchmarks \cite{Klare_2015_17419, Whitelam_2017_17829, Maze_2018_180930} with new protocols have been proposed. 
Unlike other face-related tasks such as detection \cite{Shi_2018_180923}, alignment \cite{Xu_2016_16539, Xu_2017_17394}, reconstruction \cite{Xu_2019_180917}, and soft-biometrics \cite{Sarafianos_2018_180921}, FR tasks can be evaluated with various metrics and protocols for the different scenarios.
Some datasets \cite{Huang_2007_180918, Sengupta_2016_17835} designed for face verification task are using a verification protocol comparing Receiver Operating Characteristic (ROC) curve or Precision-recall (PR) curve.
A closed-set dataset named UHDB31 \cite{Le_2017_180921} is collected and designed for evaluating FR performance in the presence of variations of pose and illuminations.
Recently, the IJB-series datasets \cite{Klare_2015_17419, Whitelam_2017_17829, Maze_2018_180930} has become available for open-set-based FR by adding more identities and variations.
As opposed to closed-set protocol, open-set FR protocols contain the novel identities in the probe but not in the gallery set in the face identification scenario.
ROC curves are reported in the 1:1 verification protocols while Cumulative Match Characteristic (CMC) curves and Decision/Identification Error Trade-Off (DET/IET) curves are reported in the 1:N mixed recognition protocols for both still images and images captured from video.

\noindent \textbf{FR Systems and Evaluation}: OpenBR \cite{Klontz_2013_14723} is a well known open-sourced computer vision and pattern recognition library in the biometrics community, which contains several FR algorithms.
However, these algorithms only work well on frontal controlled faces and the evaluation provided in OpenBR is not user-friendly.
Bob \cite{Anjos_2012_190717, Anjos_2017_090117} is another toolbox that aims to reproduce the research in signal processing and machine learning, but is difficult to install due to a large number of dependencies.
A C++ based 3D-aided 2D FR was proposed by Xu \etal \cite{Xu_2017_17643}, which used an estimated 3D model to frontalize the 2D images and matched the templates generated from the local patches according to the occlusion masks, significantly improving the performance on face images with pose variations.
However, that work only focuses on the facial template enrollment and matching stages, which does not provide a systematic evaluation package.

In summary, all these works either only provide the pre-trained model or contain some evaluation functions, which cannot be generalized to other face recognition scenarios for other researchers use.

\begin{figure*}[htb]
\begin{minipage}[b]{.58\linewidth}
  \centering
  \centerline{\includegraphics[width=\linewidth]{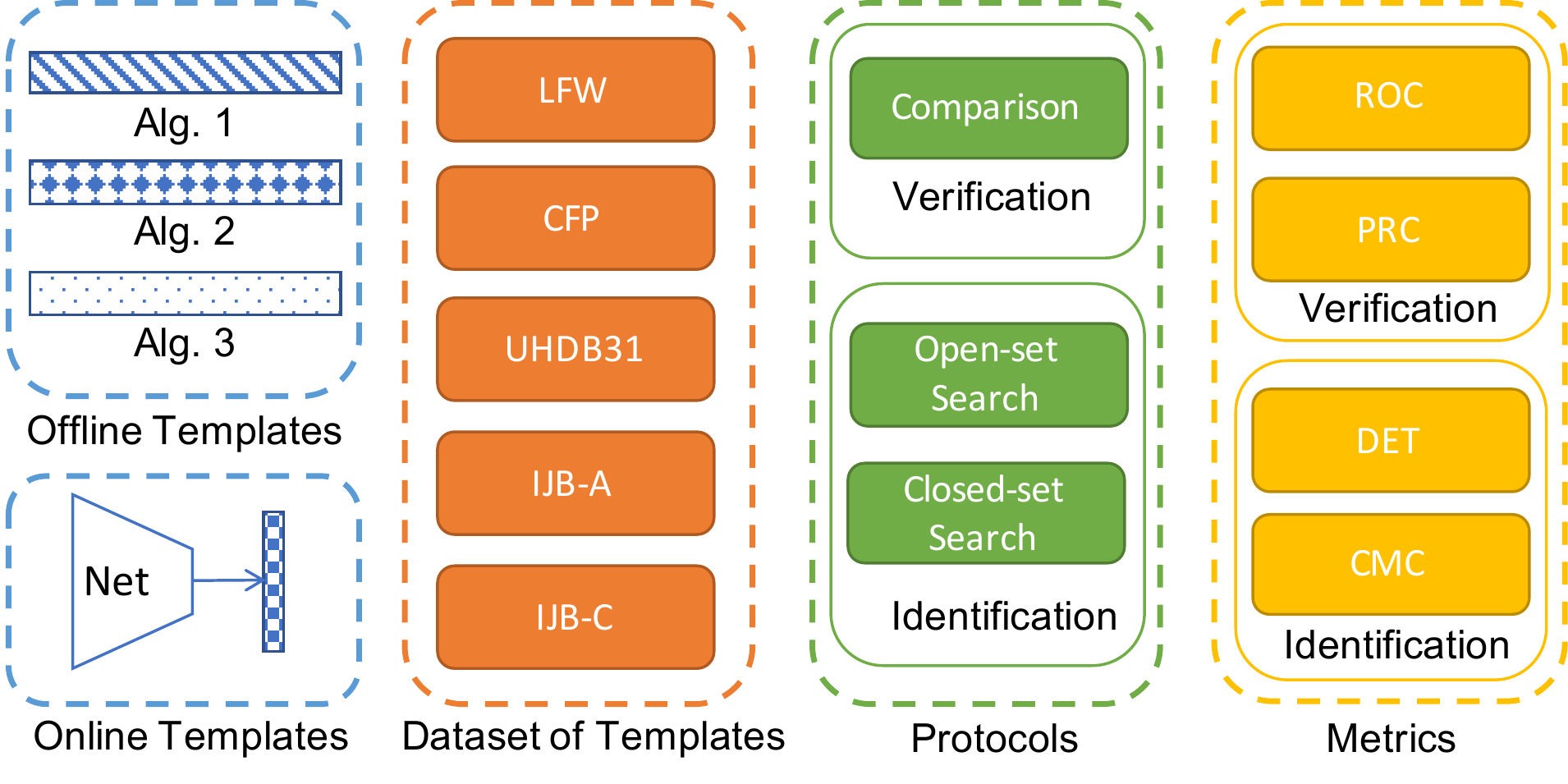}}
  \centerline{(a)}\medskip
\end{minipage}
\hfill
\begin{minipage}[b]{0.39\linewidth}
  \centering
  \centerline{\includegraphics[width=\linewidth]{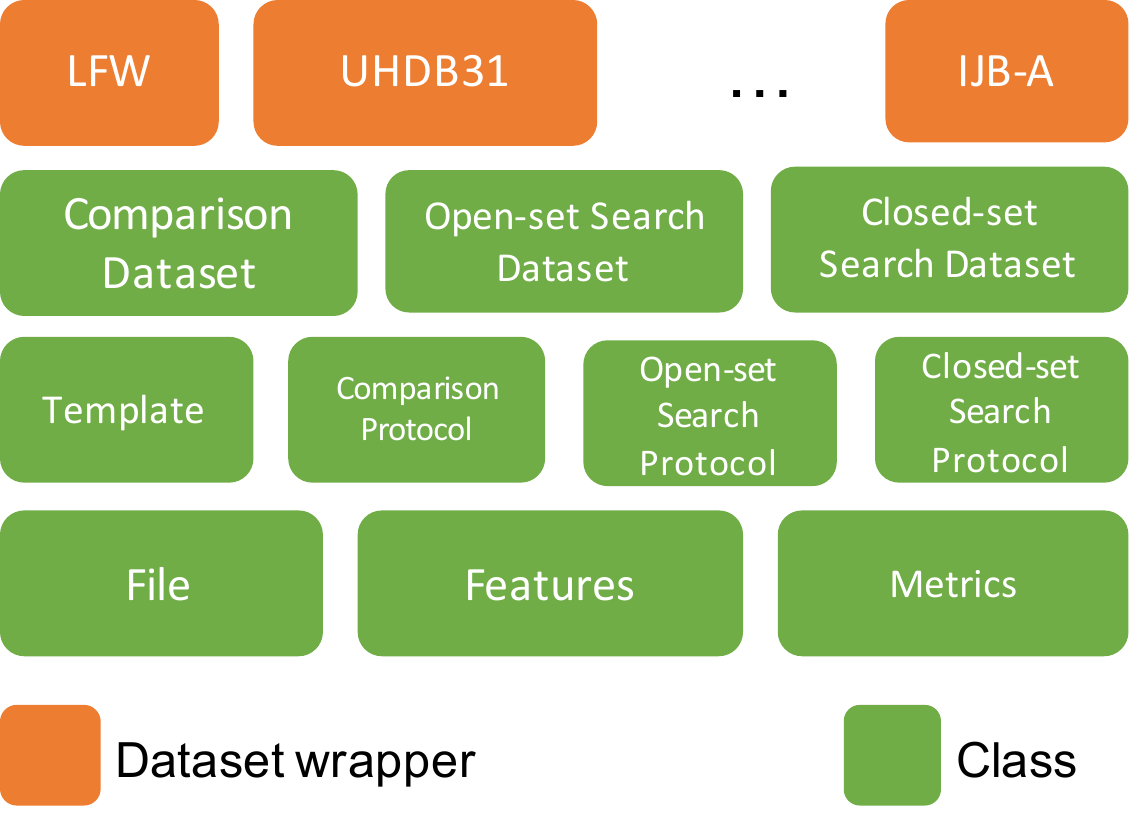}}
  \centerline{(b)}\medskip
\end{minipage}
\vspace{-1em}
\caption{(a) Depiction of different main component modules in FaRE: To support both offline and online evaluation, the templates and metadata should be mapped according to the datasets' protocol. The system can compute the similarity matrix and generate quantitative results according to the corresponding protocol used in the dataset. (b) Depiction of the system architecture of FaRE. The system is implemented in a bottom-to-up manner. The details can be found in Sec.~\ref{SEC:FaRE-Method}.}
\label{FIG:FaRE-Arch}
\end{figure*}

\section{Face Recognition Evaluation Package}
\label{SEC:FaRE-Method}
To help researchers obtain quick feedback from evaluation and accelerate research process, we design and implement a \textit{light-weight}, \textit{extendable}, \textit{generalizable}, \textit{scalable}, and \textit{maintainable} face recognition evaluation package, which can be easily generalized to evaluate other biometrics applications.
In this section, we first present a quick overview of the design and implementation of FaRE package.
Then, we describe and analyze each main component module in this package and the advantages in detail.

\noindent \textbf{Overview}:
Figure~\ref{FIG:FaRE-Arch} illustrates the main component modules, functionality, and the designed system architecture of FaRE.
Considering the generalization for commonly-used face verification datasets such as LFW and CFP and face identification datasets such as IJB-A, IJB-B, and IJB-C, two main protocols are defined in these datasets: a comparison protocol for face verification and a search protocol for face identification.
We abstract the protocols into three parts: comparison protocol, closed-set protocol, and open-set protocol.
Each protocol calls their intrinsic metric functionality to measure FR performance.
The datasets consist of the generated templates from online training or loaded templates in offline mode and the corresponding labels.
Therefore, a custom dataset can be easily extended by inheriting the existing dataset, which mainly requires to feed the templates and labels into the system.
In addition, to fit the set-based FR, a template is defined and a custom template fusion function can be easily added to generate one template from a set of feature vectors.
As depicted in Fig.~\ref{FIG:FaRE-Arch}~(b), to organize the files and templates in datasets, some classes are defined for managing the data or meta-data.
On top of the system, the users can easily call the dataset wrapper and perform the evaluation.

\noindent \textbf{Metrics}:
As one of the basic functions defined in \fare, metrics class defines and manages several commonly used metrics including ROC, PR, Accuracy (ACC), and Equal Error Rate (EER) for face verification comparison protocol, CMC and DET/IET for face identification search protocol.
These functions perform the main job of computing the quantitative metrics for researchers, which summarized in Table~\ref{TAB:FaRE-Protocol}.

\begin{table}[htb]
\begin{center}
\begin{adjustbox}{max width=\linewidth}
\begin{tabular}{lc}
    \toprule
    Protocol & Implemented Metrics \\
    \midrule
    Verification / Compare & ROC, PR, ACC, EER, AUC \\
    Closed-set Identification / Search & CMC \\
    Open-set Identification / Search & CMC, DET/IET \\
    \bottomrule
\end{tabular}
\end{adjustbox}
\end{center}
\vspace{-1em}
\caption{Summary of the current supported metrics in FaRE}
\vspace{-1em}
\label{TAB:FaRE-Protocol}
\end{table}

\noindent \textbf{Protocols}:
With pre-defined metric functions, in the comparison protocol, the system considers the ground-truth labels and the similarity vectors as input while the system requires the ground-truth labels and similarity matrix in the search protocol.
The search protocol using includes both the closed-set protocol and an open-set protocol.
In the closed-set protocol, the identities in the probe are assumed to be within the identities set in the gallery, forcing the system to assign a label from the gallery to the testing probe according to similarity ranking.
In the open-set protocol, the identities in the probe might be out of the range of identities in the gallery, which allows the system the ability to reject some samples based on their similarity scores and defined threshold.

\noindent \textbf{Datasets}:
Some dataset APIs are implemented and provided for users to quickly evaluate their algorithm on common-used datasets (LFW, CFP, IJB-A, IJB-B, and IJB-C) with different purposes.
Each dataset supports both offline and online evaluation: 
In the offline mode, the dataset will load the features from the disk and compute the similarities.
To evaluate the training process of the deep neural network, several data loaders are implemented to load the image data and forward them to the trained network to obtain the templates.

\noindent \textbf{Light-weight}:
Unlike other libraries such as Bob \cite{Anjos_2012_190717}, our package is implemented in Python and only requires a few basic dependencies such as numpy for array operation, matplotlib for visualization, scikit-learn \cite{Pedregosa_2011_190117} for metric computing, and MXNet \cite{Chen_2015_181026} for deep learning. 
Therefore, FaRE is a light-weight package because it only requires few dependencies and it is implemented by a limited number of codes, which makes FaRE extremely easy to install.

\noindent \textbf{Extensibility}:
FaRE features four extensibility aspects: adding new template fusion functions, new metrics, new protocols, and new datasets.
In set-based FR, a common way is to compute the mean feature vectors or assign different weights to compute weighted average feature vectors as the template for a set, which are implemented in FaRE.
In addition, it supports adding new template fusion functions to fuse the features from a set of images and new metrics functions to compute new quantitative measurement.
Extending current protocols or datasets is suggested to inherit the corresponding super-class and adjust the protocol process based on customized requirements, which can be quickly extended.

\noindent \textbf{Generalization ability}:
The generalization ability we define here is that the system can incorporate to different datasets, running modes, and template generators.
The package is abstracted to fit the requirements of various datasets and template generators.
FaRE supports both online and offline performance evaluation.
The online evaluation mode can be used in validating the training process while the offline evaluation mode is designed for evaluating existing algorithms.

\noindent \textbf{Scalability}:
The system can process one image or a set of images at the same time.
Several data loaders are designed and implemented to process a batch of images at the same time for online evaluation.
The researchers have options to use multiple CPUs and GPUs for evaluation in this package.

\noindent \textbf{Maintenance}:
Due to the separation of different modules and implemented logger in FaRE, the system can easily track errors and help the developer to quickly update this package. 

\begin{figure*}[htb]
\begin{minipage}[b]{.33\linewidth}
  \centering
  \centerline{\includegraphics[width=\linewidth]{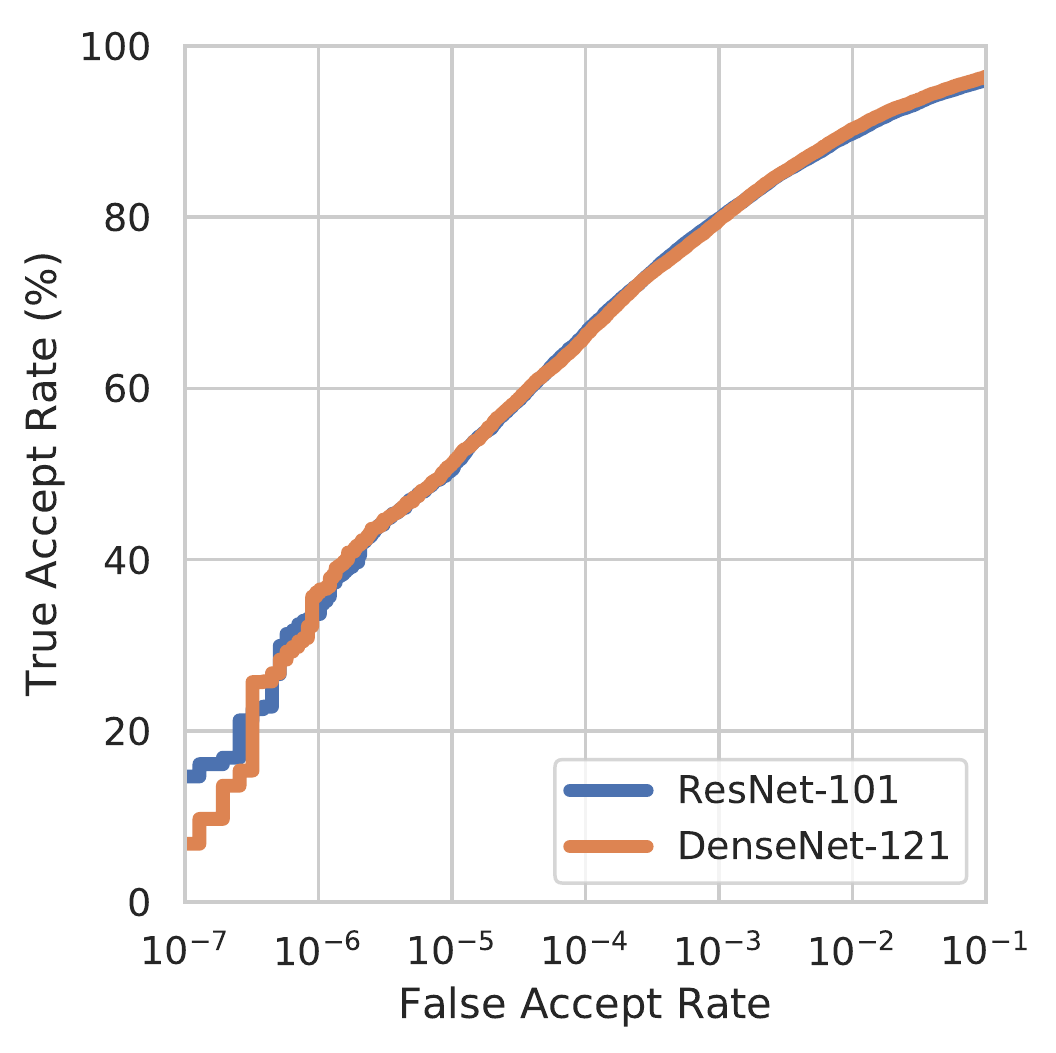}}
  \centerline{(a)}\medskip
\end{minipage}
\hfill
\begin{minipage}[b]{0.33\linewidth}
  \centering
  \centerline{\includegraphics[width=\linewidth]{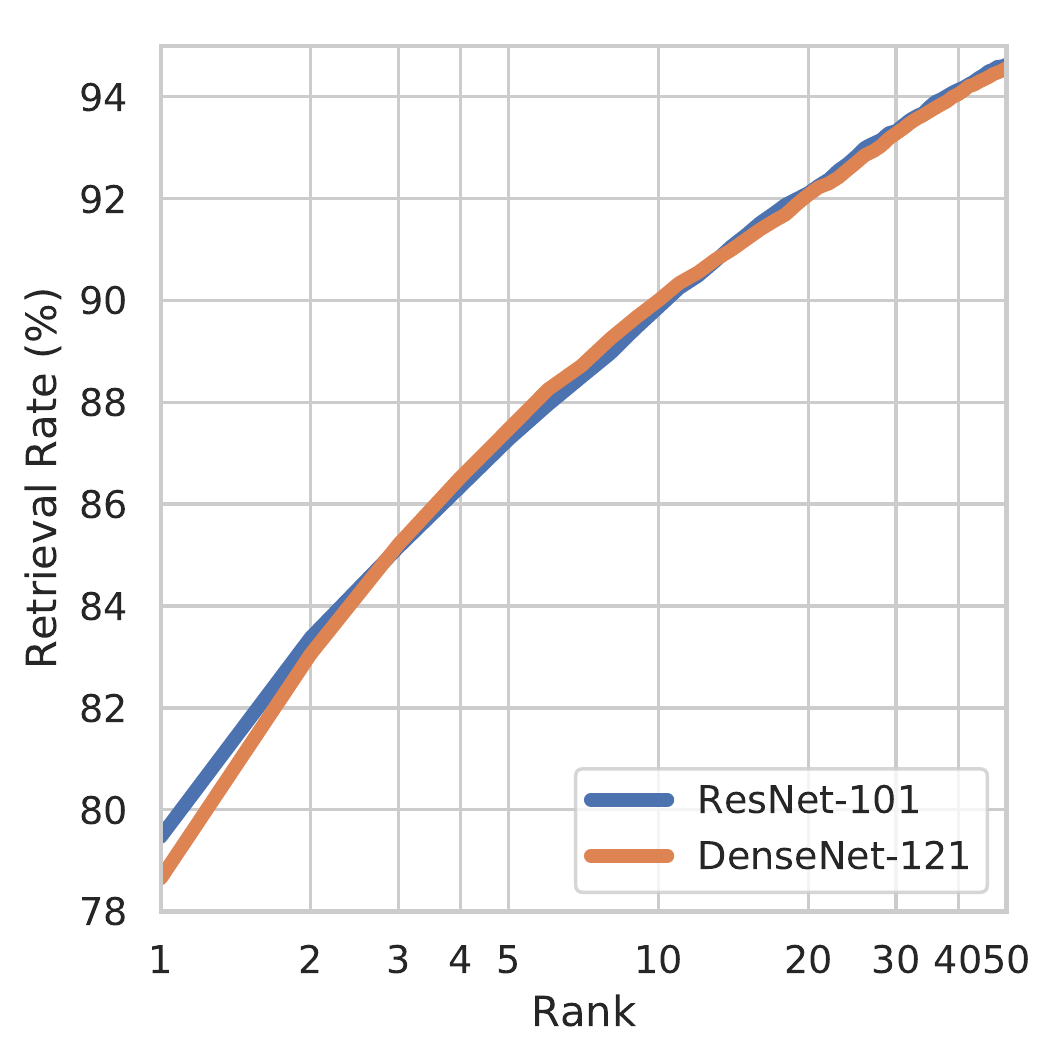}}
  \centerline{(b)}\medskip
\end{minipage}
\begin{minipage}[b]{0.33\linewidth}
  \centering
  \centerline{\includegraphics[width=\linewidth]{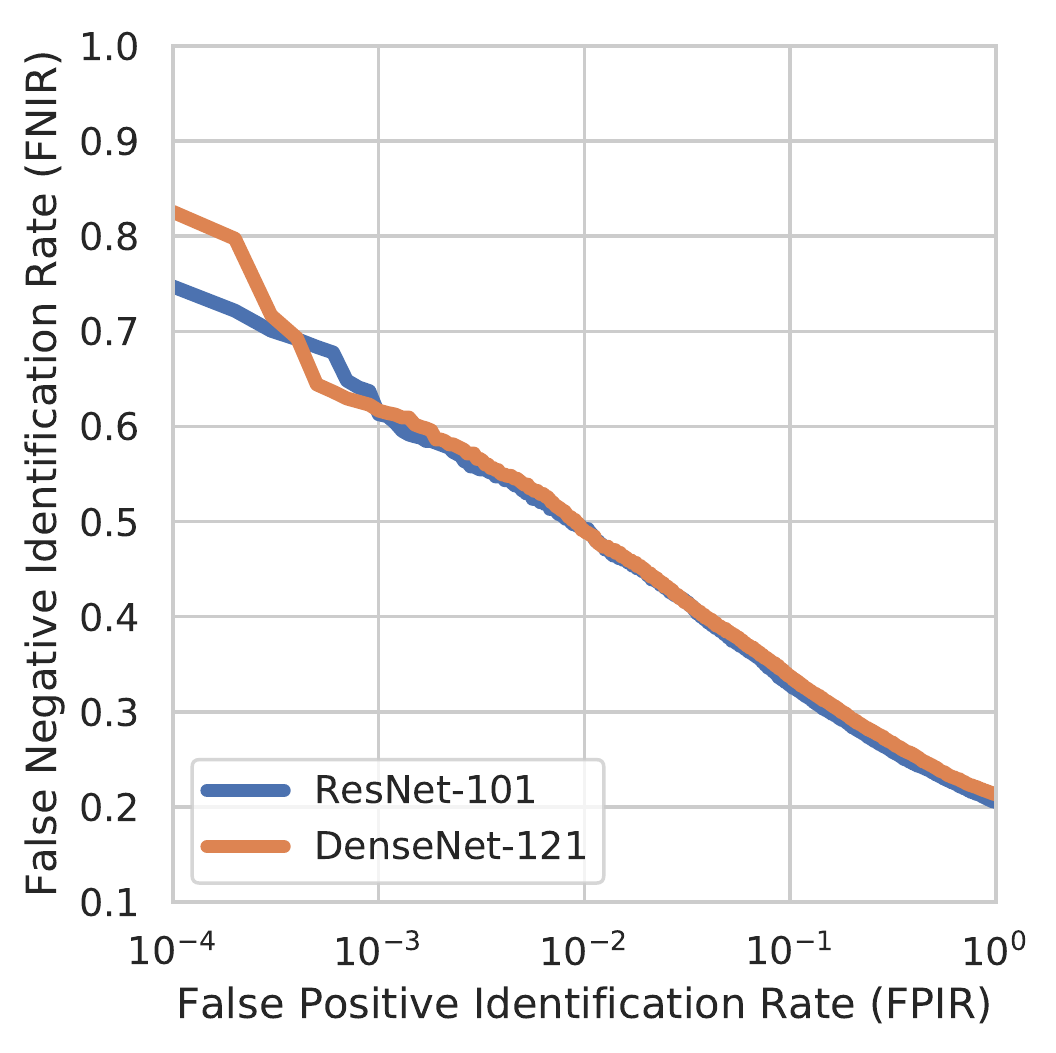}}
  \centerline{(c)}\medskip
\end{minipage}
\vspace{-2em}
\caption{Depiction of (a) the average ROC curve for the 1:1 face verification protocol, (b) the average CMC curve and IET curve for the 1:N open-set face identification protocol on the IJB-C dataset.}
\label{FIG:FaRE-IJB-C}
\end{figure*}

\section{Experimental Evaluation}
\label{SEC:FaRE-EXP}
In this section, we train two baseline models and evaluate them using our proposed package to perform the quantitative measurement and generate comparison plots.

VGG-Face \cite{Parkhi_2015_16638} is a well-known FR template generator and is provided the public access to the pre-trained model.
ResNet-101 \cite{He_2016_17161} and DenseNet-121 \cite{Huang_2017_190102} are also two famous network architectures for the object classification task.
However, there are no public pre-trained models of these two networks for the FR task.
Therefore, two baselines using ResNet-101 and DenseNet-121 are trained on VGG-Face2 dataset to generate a facial template from a single image or a set of images.
The average of the feature representations generated from a set of images of a subject is computed and treated as the template of that subject.
We evaluate FaRE using two baselines on the IJB-C \cite{Maze_2018_180930} dataset for both face verification and identification tasks to present the advantages in additional two aspects: generalization ability and scalability.
The models are trained on a GPU cluster and all evaluation experiments are performed on a local machine with a CPU of Intel Core i7-6700K and a GPU of GeForce GTX 1080-Ti.
In the online evaluation mode, it takes around one hour to generate the templates and compute the similarity scores for the mix identification task with a two-fold evaluation according to the IJB-C protocol.

To demonstrate the generalization ability, here, we set the ResNet-101 to use offline mode while DenseNet-101 is using online mode.
The mean feature vectors are computed from the set of features as the final facial template.
The average ROC performance across gallery sets for 1:1 mixed verification protocol and average CMC and IET performance across gallery sets for 1:N mixed identification protocol are computed by FaRE.
In addition, the corresponding figures are generated by FaRE as depicted in Fig.~\ref{FIG:FaRE-IJB-C}.

\begin{figure}[htb]
    \centering
    \centerline{\includegraphics[width=\linewidth]{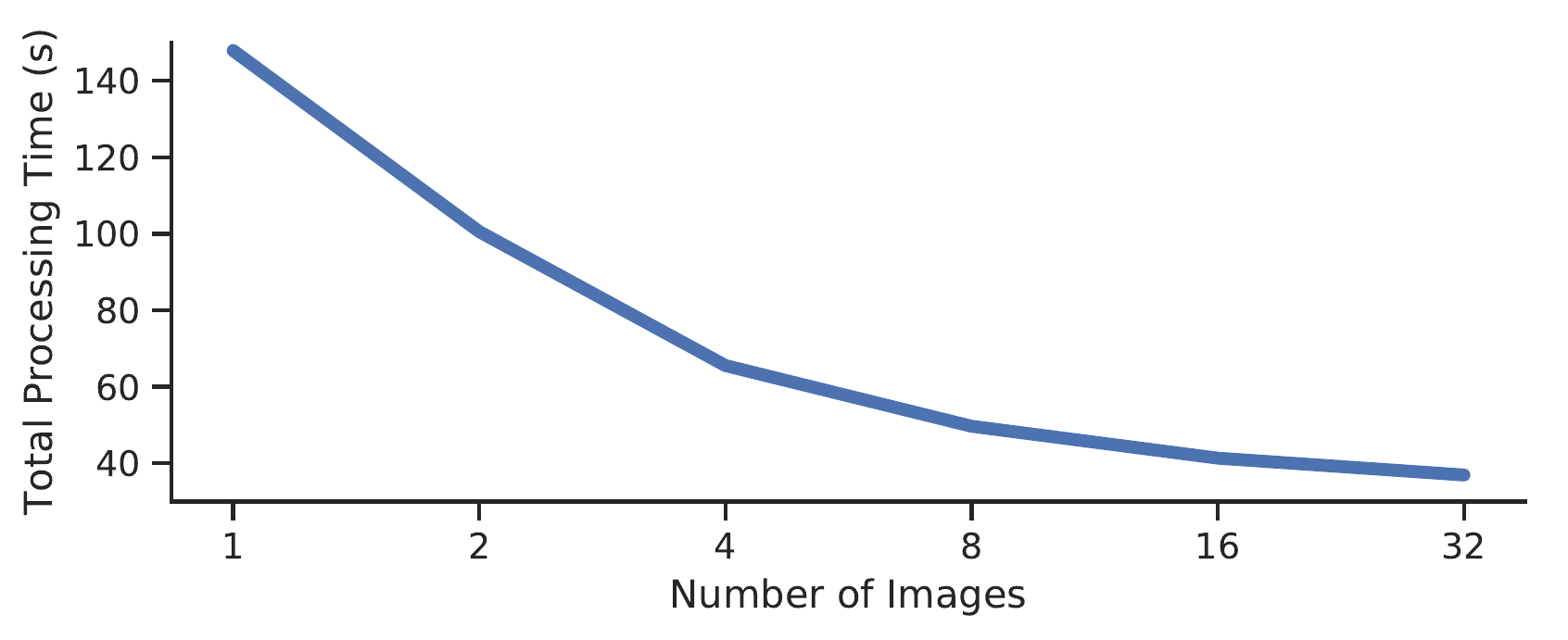}}
    \vspace{-1em}
    \caption{Depiction the scalability of FaRE.}
    \vspace{-0.5em}    
    \label{FIG:FaRE-IM-TIME}
\end{figure}

To demonstrate the scalability, for simplicity, we directly performed 10-fold evaluation using FaRE on LFW dataset in the online evaluation mode.
The relation of a number of images processed at the same time with the total processing time of generating the templates and computing the similarity scores is depicted in Fig.~\ref{FIG:FaRE-IM-TIME}.
It takes approximately 35 seconds to finish generating templates using DenseNet-121 and comparing all pairs in LFW by processing 32 images at the same time, which can provide quick feedback for algorithm development stage.

\section{Conclusions}
\label{SEC:FaRE-Cons}
In this work, we designed and implemented a light-weight, maintainable, scalable, generalizable, and extendable face recognition evaluation toolbox in Python named FaRE that supports both online and offline evaluation to benefit the biometrics research community and to accelerate the biometrics-related research.
FaRE is designed to evaluate general FR systems, which consists of current commonly-used evaluation metrics functions, closed-set, and open-set FR datasets, and can be extended to other customized datasets. 
Two baselines are evaluated on the IJB-C datasets to provide baselines and pre-trained model for other face-related tasks.

\smallbreak
\small{
\noindent \textbf{Acknowledgment}
This material is supported by the U.S. Department of Homeland Security under Grant Award Number 2017-ST-BTI-0001-0201 with resources provided by the Core facility for Advanced Computing and Data Science at the University of Houston.
}

\pagebreak
{
\small
\bibliographystyle{IEEEbib}
\bibliography{refs}
}

\end{document}